\documentclass{elektr}
\RequirePackage{fix-cm}
\usepackage{hyperref}
\hypersetup{
colorlinks=true,
urlcolor=blue,
citecolor=blue}
\usepackage[all]{xy}
\usepackage{amsfonts,amssymb,amsmath,amsgen,amsopn,amsbsy,theorem,graphicx,epsfig}
\usepackage{amscd,bezier,latexsym,mathrsfs,enumerate}\usepackage[utf8]{inputenc}\usepackage[english]{babel}
\usepackage[dvipsnames]{xcolor}
\usepackage[pagewise]{lineno}
\usepackage{multirow}
\usepackage{tabularx}
\usepackage{ragged2e}
\usepackage{blindtext}

\year{}
\vol{}
\fpage{}
\lpage{}
\doi{}

\title{Real-Time Trash Detection for Modern Societies using CCTV to Identifying Trash by utilizing Deep Convolutional Neural Network}

\author[AUTHOR and AUTHOR]{
\textbf{Syed Muhammad Raza$^{1}$, Syed Muhammad Ghazi Hassan$^{1,2}$, Syed Ali Hassan$^{1,3}$, Soo Young Shin$^{1,4}$\thanks{elektrik@tubitak.gov.tr}~}\\
$^{1,2}$Department of Computer Science, PAF-Karachi Institute of Economics and Technology, Pakistan\\
$^{3,4}$Department of IT Convergence, Kumoh National Institute of Technology, South Korea\\

\\ [1.8em]

\rec{.201}
\acc{.201}
\finv{..201}
}

\def\E{\ifmmode{\mathbb E}\else{$\mathbb E$}\fi} 
\def\N{\ifmmode{\mathbb N}\else{$\mathbb N$}\fi} 
\def\R{\ifmmode{\mathbb R}\else{$\mathbb R$}\fi} 
\def\Q{\ifmmode{\mathbb Q}\else{$\mathbb Q$}\fi} 
\def\C{\ifmmode{\mathbb C}\else{$\mathbb C$}\fi} 
\def\H{\ifmmode{\mathbb H}\else{$\mathbb H$}\fi} 
\def\Z{\ifmmode{\mathbb Z}\else{$\mathbb Z$}\fi} 
\def\P{\ifmmode{\mathbb P}\else{$\mathbb P$}\fi} 
\def\T{\ifmmode{\mathbb T}\else{$\mathbb T$}\fi} 
\def\SS{\ifmmode{\mathbb S}\else{$\mathbb S$}\fi} 
\def\DD{\ifmmode{\mathbb D}\else{$\mathbb D$}\fi} 

\newcommand{\bse}{\begin{subequations}}
\newcommand{\ese}{\end{subequations}}
\newcommand{\ben}{\begin{enumerate}}
\newcommand{\een}{\end{enumerate}}
\newcommand{\bens}{\begin{enumerate*}}
\newcommand{\eens}{\end{enumerate*}}
\newcommand{\be}{\begin{equation}}
\newcommand{\ee}{\end{equation}}
\newcommand{\bea}{\begin{eqnarray}}
\newcommand{\eea}{\end{eqnarray}}
\newcommand{\baa}{\begin{eqnarray*}}
\newcommand{\eaa}{\end{eqnarray*}}
\newcommand{\bc}{\begin{center}}
\newcommand{\ec}{\end{center}}

\theoremstyle{corollary}

\theoremstyle{lemma}

\theoremstyle{proposition}

\theoremstyle{axiom}

\theoremstyle{conjecture}

\theoremstyle{example}

\theoremstyle{definition}

\theoremstyle{remark}

\begin{document}

\maketitle

\begin{abstract}To protect the environment from trash pollution, especially in societies, and to take strict action against the red-handed people who throws the trash. As modern societies are developing and these societies need a modern solution to make the environment clean. Artificial intelligence (AI) evolution, especially in Deep Learning, gives an excellent opportunity to develop real-time trash detection using CCTV cameras. The inclusion of this project is real-time trash detection using a deep model of Convolutional Neural Network (CNN). It is used to obtain eight classes mask, tissue papers, shoppers, boxes, automobile parts, pampers, bottles, and juices boxes. After detecting the trash, the camera records the video of that person for ten seconds who throw trash in society. The challenging part of this paper is preparing a complex custom dataset that took too much time. The dataset consists of more than 2100 images. The CNN model was created, labeled, and trained. The detection time accuracy and average mean precision (mAP) benchmark both models' performance. In experimental phase the mAP performance  and accuracy of the improved CNN model was superior in all aspects. The model is used on a CCTV camera to detect trash in real-time.

\end{abstract}

\section{Introduction}
\label{Int}
As the city of lights, Karachi is being listed as the dirtiest city in the world according to global air quality. The population is increasing gradually, and the land areas are dividing into societies. Everyone wants to live in an area that is not polluted. According to the Air Quality Index, Karachi city is the second most polluted city in the world. The amount of contaminated particles in the city's air has been recorded at 274 particulate meters. From international news, Karachi, formerly known as the city of lights, is now the most polluted city, with a pile of rubbish and canals everywhere. According to authorities, the city has 250,000 tons of leftover garbage dumped in residential areas. The city of Karachi produces daily up to 16,000 M / tons of domestic waste alone, commercial and industrial added to this, and the daily disposal rates at landfills are between 7,000 and 8,000 tons. Throughout the year, we see that garbage dumping in Karachi has become one of the biggest urban planning problems. As population, urbanization, and industry increased, garbage and pollution throughout the city grew exponentially. It is not easy to find a large piece of land or a portion of clean land in Karachi. Garbage includes all types of waste, including household waste, construction waste, and material waste. Small mounds of rotten rubbish are found near food markets, and plastic shopping bags are scattered everywhere. Plastic bags are a big problem as they are not flexible. Karachi is still the city's economic and financial center but has changed from an urban Pakistani city to a garbage city. Tons and tons of garbage are dumped on street corners, behind mosques, in closed schools, and where possible. The significant challenges are urban sprawl. The biggest challenge is the increase in waste produced and garbage disposal due to the high demand for food products and other products. Public garbage bins are being filled faster, and many containers are likely to overflow, creating overcrowded roads and bad smells, and adverse health and environmental effects. According to UNEP, more than 400 M tons of garbage are produced every year. Most of the trash is made in industrial areas, and this garbage does not dump properly, and authorities do not pick most of the garbage.

But most of the area is highly polluted and does not meet the criteria of living standards. Instead of dustbins, people throw trash wherever they want. Modern societies are developing and making life more comfortable with luxury standards. In modern societies, the authority installs smart CCTV cameras on every street, and their priority is to maintain and make the society clean and green. 

Garbage disposal is essential and related to the fact waste can be treated in a rounded and safe manner. The classification of operation and treatment is not suitable for long-term screening due to the high temperature and humidity conditions associated with the presence of toxic gases and odors, garbage [1].

The subfield of Machine Learning is Deep Learning (DL). It is involved in the functioning and formation of the brain structure known as the AI neural network. When it comes to text recognition and especially fonts from images  DL plays a significant role. The fonts and recognition of the famous letter (Turkish) are made utilizing the DL technique. Brain cancer and other diseases can be classified by using deep learning. [2]. 
The interest of researchers is increasing gradually in CNN because detection is an automatic crucial during the learning stage. The excellent performance was shown by the CNN model in image classification and object detection. The main reason behind the superb performance is a CNN model can extract the abstract features [2]. A convolutional neural network was used for recognition, detection, processing, and segmentation. Moreover, the performance of CNN is gradually upgrading because it extracts the images automatically. In recent years, the advancement shows that 
various CNN object detectors have been introduced in CNN-based object detection, e.g., Faster R-CNN and SSD. In particular, the CNN-based object two most important types of detectors are: (1) single image and (2) region-based. A powerful GPU is required to function for these detectors because they are computationally intensive, while the single-image detectors consist of a convolution network for detection. SSD is a kind of single shot detector [3]. The operation of the classification CNN model is very much more accessible. To comprehensive understanding convolutional neural networks, in CNN, the first initial layer recognize most the straightforward structures of the image,for instance, with standard image of two-dimensional.To identifies one of the edges in images each filter path is mandatory. Therefore, when the edge details included in the image are taken into account, different filters work at different angles at the edge. To detect filters on the first layer of 
output, the features maps contain the information structures of filter. The images are associated with various edge structures, which include the information in the results. The previous layer on feature maps the edges that have been detected reveals on the next evolution layer. Each convolutional layer examined the correlations of combination structures detected in the previous layer with images. In this process, the number of layers increases with complexity, facilitating semantic information from structural information.

In the following way, the convolutional layer works. The layer gets the insertion volume. In this case, an image has a certain height, width, and depth. There are matric filters available that start with random numbers initially. The input images have the same depth as the channels, and the filters are small spatially. For Grayscale, filters will have one depth, RGB filters will have three depths, and so on. The filter is convolved over input volume. The filter is displayed above the input volume.  Slide spatially with an image and analyze dot products in the entire picture. Filters eventually produce maps to activate the input image.

The convolutional neural network (CNN) model with a large number of specified datasets helps detect trash through a CCTV camera and identify the trash. Simultaneously, CCTV records video for 10 seconds of the object who throw trash. The one who throws trash will be easily identified, and further actions will be taken against the specified object.

The paper pattern is organized as follows; Section II describes the research-related work. Section III describes the experimental analysis, which represents the set of data and the results obtained. Finally, section IV concluding the paper around the end. In the final phase V, future work and conclusion.

\section{RELATED WORK}
\label{Int}
In the past, none pertain specifically to trash detection. Many research projects have been done related to a convolutional neural network (CNN) based on image classification and support vector machine (SVM).
In the field of image classification, AlexNet is one of the best known and particularly effective convolutional neural network (CNN) structures. The challenge was won by (ILSVRC). Relatively the architecture is simple to understand, and it performs well. The trend of the CNN model was started by AlexNet it was an influential approach and very popular in the ImageNet, and its image classification was state of the art [4].
The subtraction of background was generally use as an old former techniques [5] and another approach of different classification techniques like Haar cascade to detect the objects [6]. One of the projects from TechCrunch Disrupt Hackathon [7] where the team builds an "Auto-Trash", automatic garbage can you see between compost and reuse using a Raspberry Pi camera and camera module. The project was make-up using Google's Tensor Flow and incorporates hardware elements. One thing to note about Auto-Trash is that it only differentiates whether something is compost or recycling, which is more accessible than getting six classes. Another way to get trash detection is to separate the garbage. Waste disposal improved the efficiency of recycling and improved the city's appearance, environmental protection, increased resources, and the same social, environmental, and economic benefits. In recent years, garbage disposal has been a significant concern for the government.
	  
The TrashNet data set of image classification is comparative analysis. In this phase, without using the data augmentation method half of the data set was used for testing purposes. A fine-tuned model such as GoogleNet, ResNet, VGG-16, SquezeeNet, and AlexNet was proposed. Softmax and SVM are two different classifiers are used for the highest classification accuracy [8]. 

The sanitation department provides a large quantity of urban garbage images dataset from a garbage dataset. Afterward, the object category dataset is obtained from the non-garbage urban scene images dataset from background classes. In the test run, there is no display for background classes, and only garbage class is shown, thus improving the algorithm robustness [9]. The application of action recognition is the behavior of detecting garbage dumping. Similarly, the method for the adoption of conventional action recognition is complex. The details of required data of the model appearance of dumping behavior using deep learning algorithms are not enough. Besides, there are several appearance variations of the trash discarding behavior. The general appearance model is complicated. [10].

Waste identification is an essential step before segregation and can best be done with the help of different machine learning and image processing algorithms. Convolutional neural networks (CNNs) are highly preferred in image classification. CNN allows us to extract distinct elements from an image and classify them into pre-determined categories [11]. As a result of the introduction of GPUs, computing power has increased dramatically, which is why large image data sets can be processed in such a short time. As a result, CNN gained popularity in the last few years [11].

The classification of plastic and non-plastic was done when a image is taken by a camera and transfer to  CNN code for pre-processing. The image collection is done in a frame using a CNN, and the CNN is made up of two hidden layers and one perfectly aligned layer that gives the effect of whether the image is plastic or non-plastic [12]. CNN develops a category of models in the broader field of artificial neural networks, and they have been shown to produce excellent results with efficiency and accuracy. Their advantages compared to traditional methods can be summarized into three main points: features are categorized to create a high-level representation, the use of deep structures increases the power of precision, and the architecture of CNN allows for the integration of related tasks [13].

The motor actions can be used by a convolutional neural network, the camera images are directly extracting the useful features [8]. The test consists of learning to the robotic from monocular images using RL. There is 800,000 capture try completed in more than 14 robotic manipulators. Many resources are required, and the learning time makes use of such a method ineffective in real-world situations. If the optimizer gives the trajectories for solving fraudulent activity, it is utilized to lead CNN policy searches to good local optima and speed up the process [14].

To locate and recognize the trash in images captured by a web camera attached to a robot, utilizing the convolutional neural network (CNN) model. [15].

Further, research of MobileNets was developed by Google researchers by using CNN's class. Like "mobile-first," because it is easy to use and runs fast on mobile phones. The two parameters of MobileNets are width multiplier and resolution multiplier, which are tuned to weigh the resource-accuracy tradeoff. The resolution multiplier changes the input image dimension, while width multiplied can thin the network. These changes can reduce the interior structure of all layers [16].

Waste reuse is given as (1) government budget and program: adequate govt. Legislation and MSW management budget; (2) home education: importance of recycling loose by household; (3) technology: The reuse technology is less effective; and (4) administrative costs: Manual separation is highly cost. Recent advances of improvements in computer vision DL have contributed to unprecedented. Convolutional neural network (CNN) is one of the DL algorithms for its extensive range of image processing, classification, and discovery. In these books, CNN proposes to create a waste classification [17].
Sakr et al's scientific research attempts to make waste filters more efficient by utilizing ML algorithms. The author used two popular methods, DL with convolutional neural networks (CNNs) and vector support devices (SVMs). The obtained results of SVMs achieved a high separation accuracy of 94.8{\%} while CNN received only 83{\%} [18].
In a sustainable society, recycling is becoming an indispensable part. Moreover, the whole recycling procedure demands massive hidden costs caused by the selection, processing, and classification of recycled materials. Garbage sorting these days in many countries, they may be confused about finding the proper garbage dump where most types of building materials are. Finding the default way is now significant for industry and knowledge society, which has environmental effects and beneficial economic outcomes. The few parameters of CNN's are easy to train. However, for further investigation, various models according to convolutional neural networks are created for waste segregation. Thus, this study confirmed one item in an image and divided it into recycling classifications like plastic, paper, and metal [19].
In crowded areas, garbage detection is very crucial for the environment. Manual monitoring is time-consuming and requires a lot of labor. With the help of airborne hyperspectral data, a method was introduced to monitor garbage distribution in crowded area. For environmental monitoring in vast areas, airborne hyperspectral remote sensing has the advantage of using it. To generate a binary garbage segmentation map and classify the pixels of HSI data, a new hyperspectral image classification network MSCNN was proposed for the garbage detection. The garbage detection was divided into two-step tasks, unsupervised object detection, and HSI classification. The dataset of HSI garbage was labeled by us since the HSI garbage dataset is not publicly available. Nowadays, in many fields, deep learning has proved its efficiency in robustness. In deep learning, many HSI classification algorithms are based. In HSI, a high-level features extractor to extract used in convolutional neural networks (CNN), the classification task was completed on multi-layer Perceptron (MLP) [20].
On-street garbage city managers spend a large amount of energy and money cleaning for the random appearance. The visual street cleanliness assessment is critical. The collection of street garbage was cleanliness information is not real-time and also not automated. However, these existing assessment approaches have clear disadvantages. To collect the street images, the first-ever high-resolution camera was installed on vehicles. To extract the street image, the Mobile edge servers are temporarily used for storing the image. Through the city network, the processed street data is transmitted to the cloud data center for analysis.
Moreover, for the mobile edge processing, two different edge servers were used to complete two tasks. The entire system performance was improved in the first task. To modify the suitable size of the pictures in this stage, the image data is sent to the convolutional neural network (CNN) as a first input when object detection is performed. Secondly, to reduce the overall time of the entire system, the image data is preprocessed in the edge server. When the image is transmitted to the edge server, a modified algorithm changes the size of the image automatically. In the mobile station, the vehicle in the city was explicitly used for garbage collection. In the garbage, collection vehicles install the high pixel, high resolution, and network transmission capabilities cameras on the vehicle's top. The distance cover by a camera in front 50 meters faces the ground. The regular images are captured, data are transmitted into the edge server on time by the garbage collection vehicle, and urban citizens also do garbage collection. The collected data is sent through the edge server from their mobile device. The edge of the network is also called the edge server. The mobile device requested to connects through wireless data links to handle a portion [21].

\section{PROPOSED SCHEME}
Keeping the concept of Modern Society and fixing the problem of trash throwing in societies, catching the people red-handed, and assuring to protect the environment from pollution by using artificial intelligence AI and the Deep CNN model. In the beginning, the proposed contribution in deep CNN model manifests affection on an improved model. Initially, the default CNN model was used to train the dataset. However, the obtained results were not satisfactory, but the model's speed is upstanding. To achieve precise model accuracy, the layers of CNN 1024 x 1024 were added to the model. As a result, the accuracy of the model is increased, and proportionally the speed is decreased. The input of the images was divided into grids cells G x G is kept as 13 x 13. Each grid cell is responsible for the probability of object in the image during the learning stage of the model. The LableImage tool was used to label the image, producing a bounding box with the help of five consisting predictions. The cell predicts a bounding box in the form of a rectangle shape. The prediction of the bounding box includes the horizontal and vertical components, at center height and width, with a confidence score considering the image in a grid cell. These components are labeled as 'X' and 'Y,' 'H' and 'W'; finally, confidence score with 'Cs.' A confidence score 'Cs'  reflects the accuracy of a model that contains a predicted object. The object in the bounding box is accurately determined by the confidence score responsible for prediction. To reducing activation function, the ReLU is utilizing, and the gradient is vanishing and achieve sparsity. For the smooth propagation of information across deeper layers, the two activation functions used ReLu and Mish. ReLU avoids the gradient from destroying, and Mish helps to prevent capping. To avoid the slow training, the first fifteen layers is used to avoid the saturation in the Mish, while unbounded below property results in a regularization effect. In the improved CNN model the  Mish activation function was replaced with leaky ReLU activation function. The generalization  and training time, the trade-off is considering finalized in number of  convolutional layers. To maintain the accuracy an extensive hit and trial, minimize the computational complexity to obtain the minor prediction error and simpler model.
\begin{center}
    \textbf{Table 1}Performance of default deep CNN model and improved CNN model
    \includegraphics[width=8.55cm]{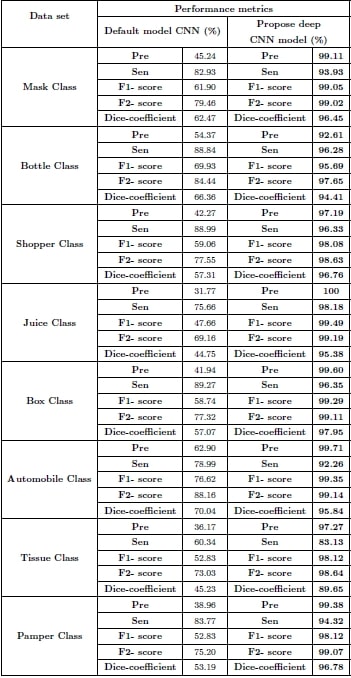}
    \includegraphics[width=8.34cm]{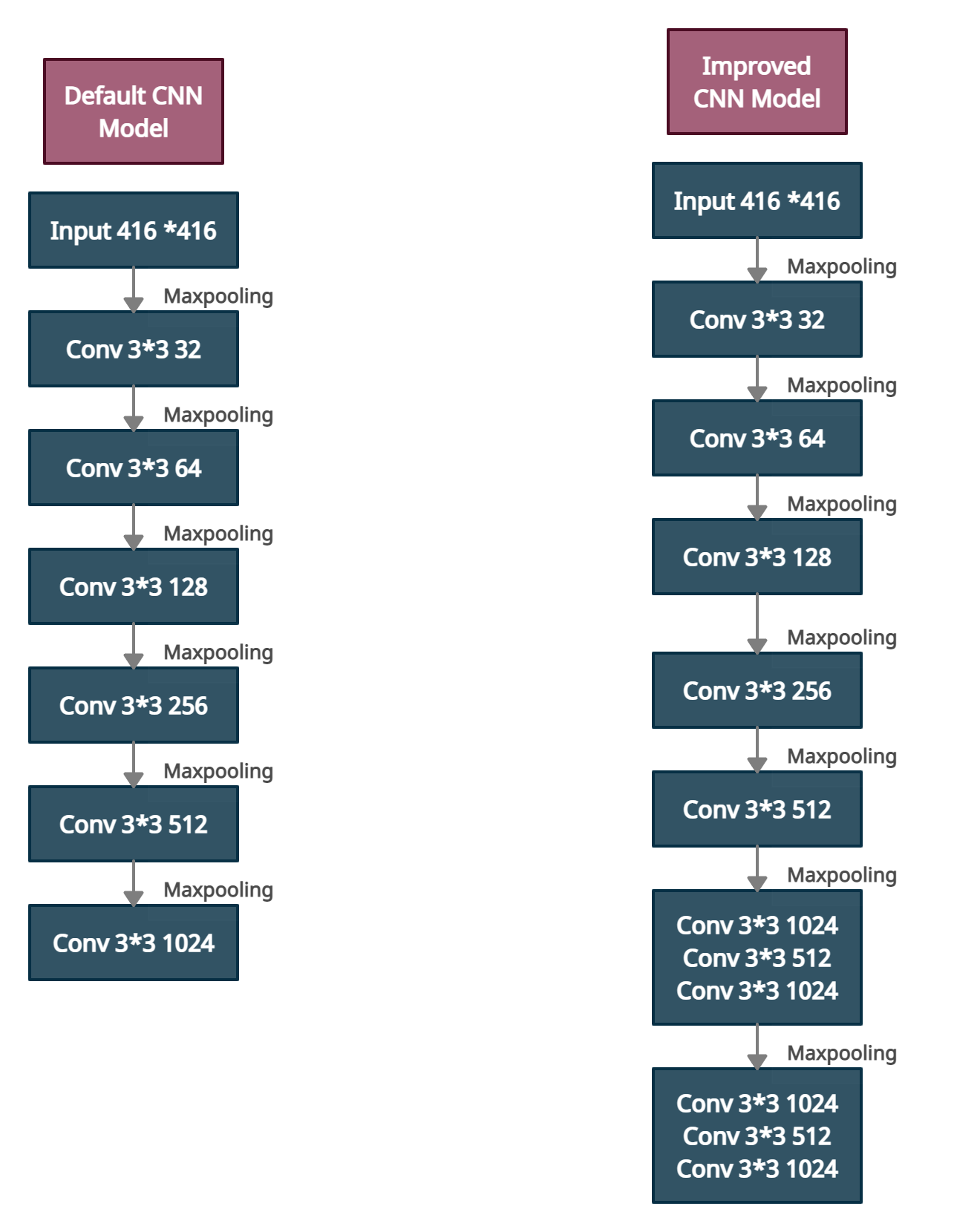} 
    \begin{FlushRight}
        Fig 01: Default and Improved CNN Model    
    \end{FlushRight}
     
\end{center}

\section{EXPERIMENTAL ANALYSIS OF WORK}
\label{Int}
This section emphasizes the functionality of the model. The obtained dataset and detail specifications of experimental results and performance are produced from this model.
\subsection{Dataset Specifications}
The process of collecting data was done manually because there are no publicly available datasets for trash detection. The entire dataset for the improved CNN model consists of 8 classes of trash with more than 2100 images, and each class consists of 300 images.  \textbf{These classes of the dataset are masks, tissue papers, shoppers, boxes, automobile parts, pampers, bottles, and juice boxes.} There is no use of data augmentation for a dataset. For this purpose, make a video of every class of trash from different angles and break the video into frames. Then a video was converted into images by utilizing Python script. The images of the dataset are given below.
 
\begin{center}
    \begin{center}
        \includegraphics[width=2.9cm]{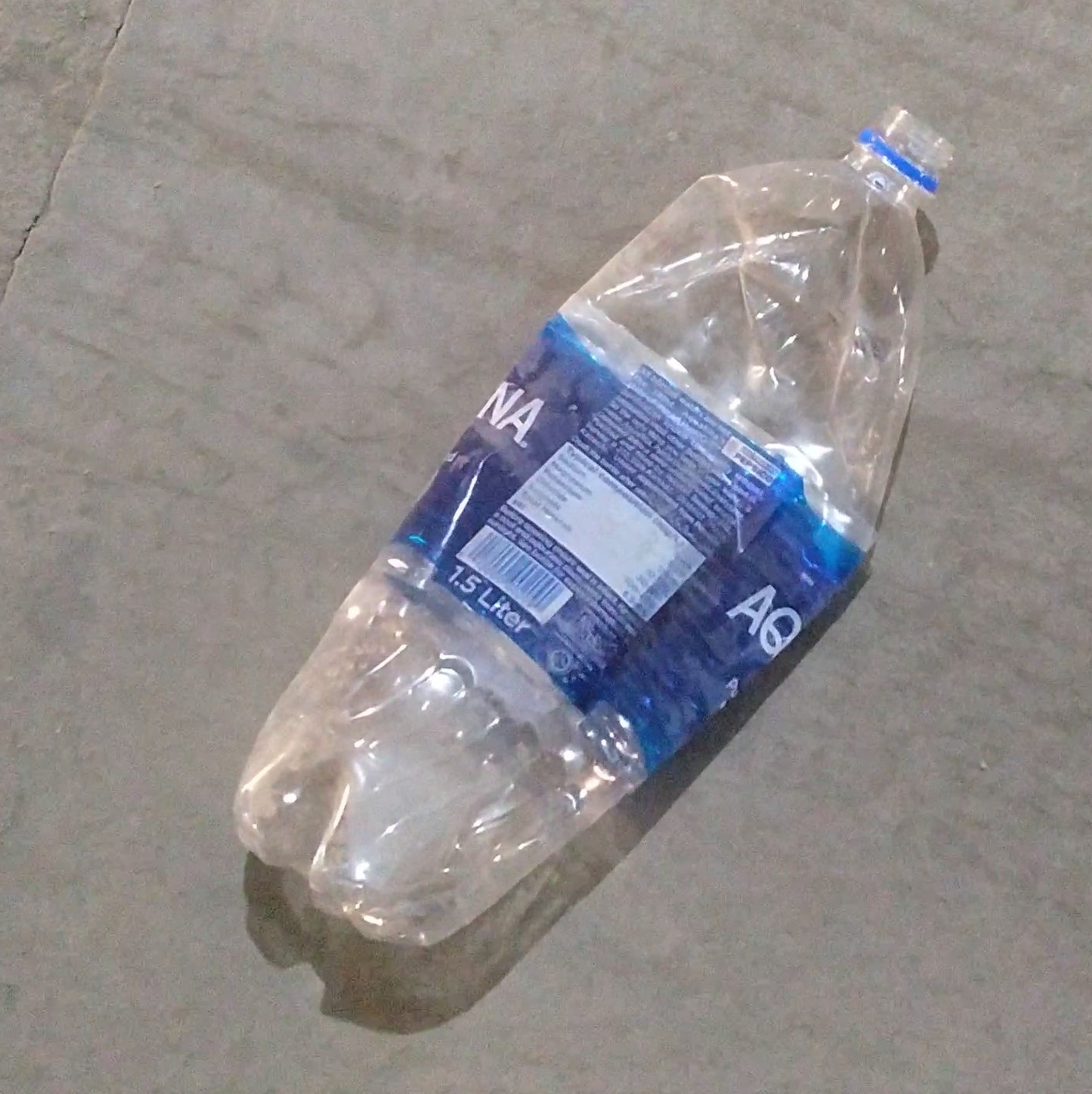}
        \includegraphics[width=2.8cm]{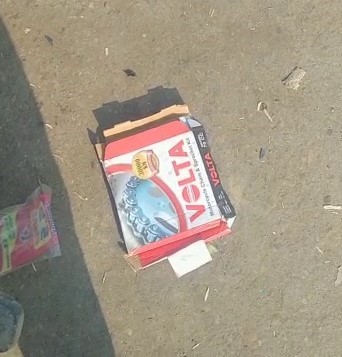}
        \includegraphics[width=3cm]{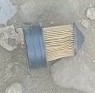}
        \includegraphics[width=2.4cm]{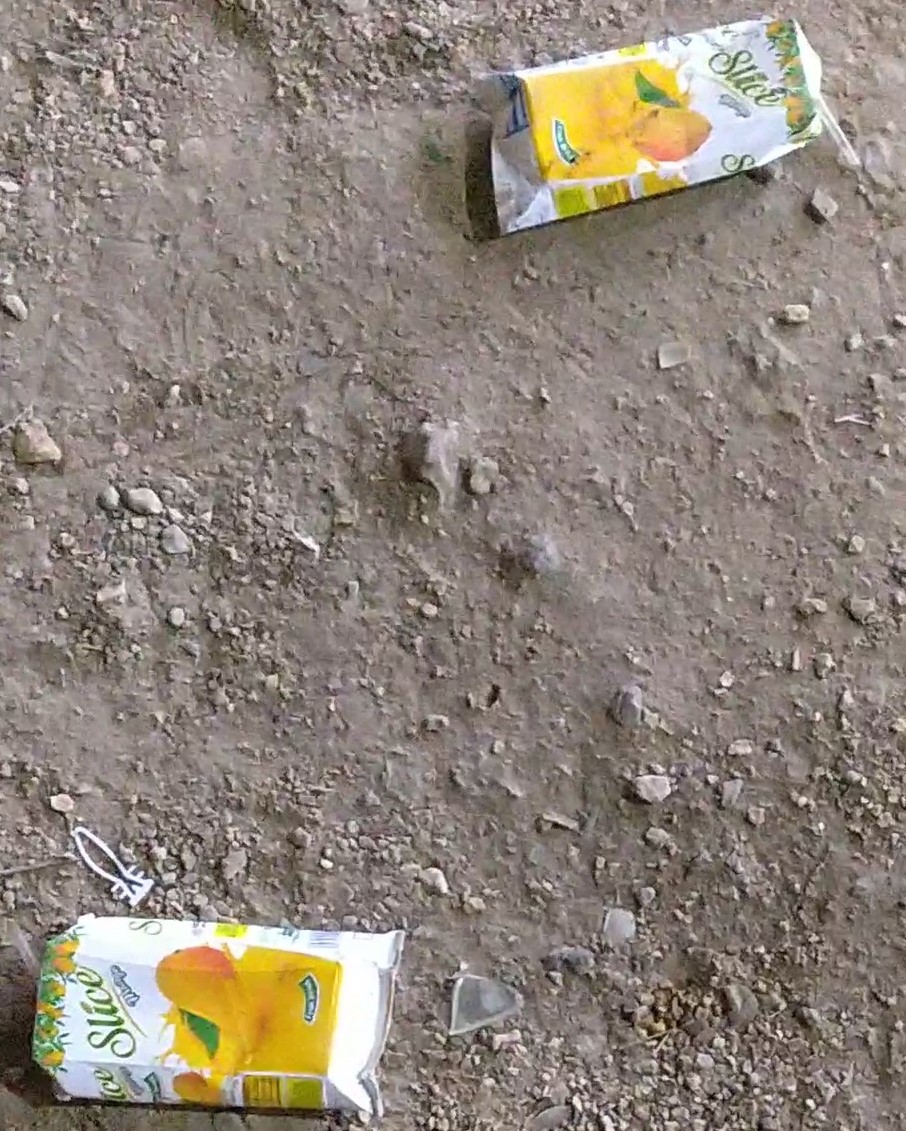}
        \newline
        \includegraphics[width=2.5cm]{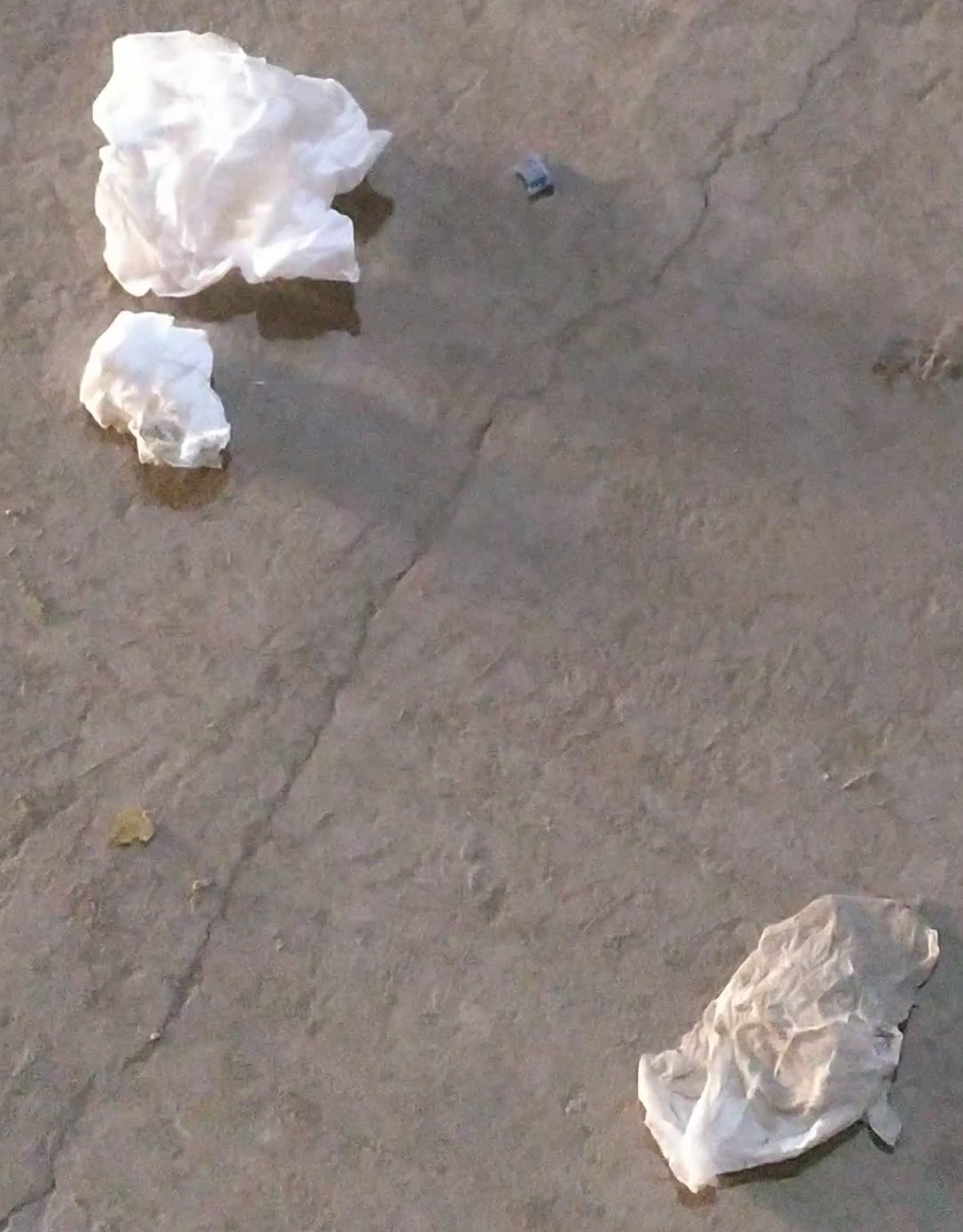}
        \includegraphics[width=4.2cm]{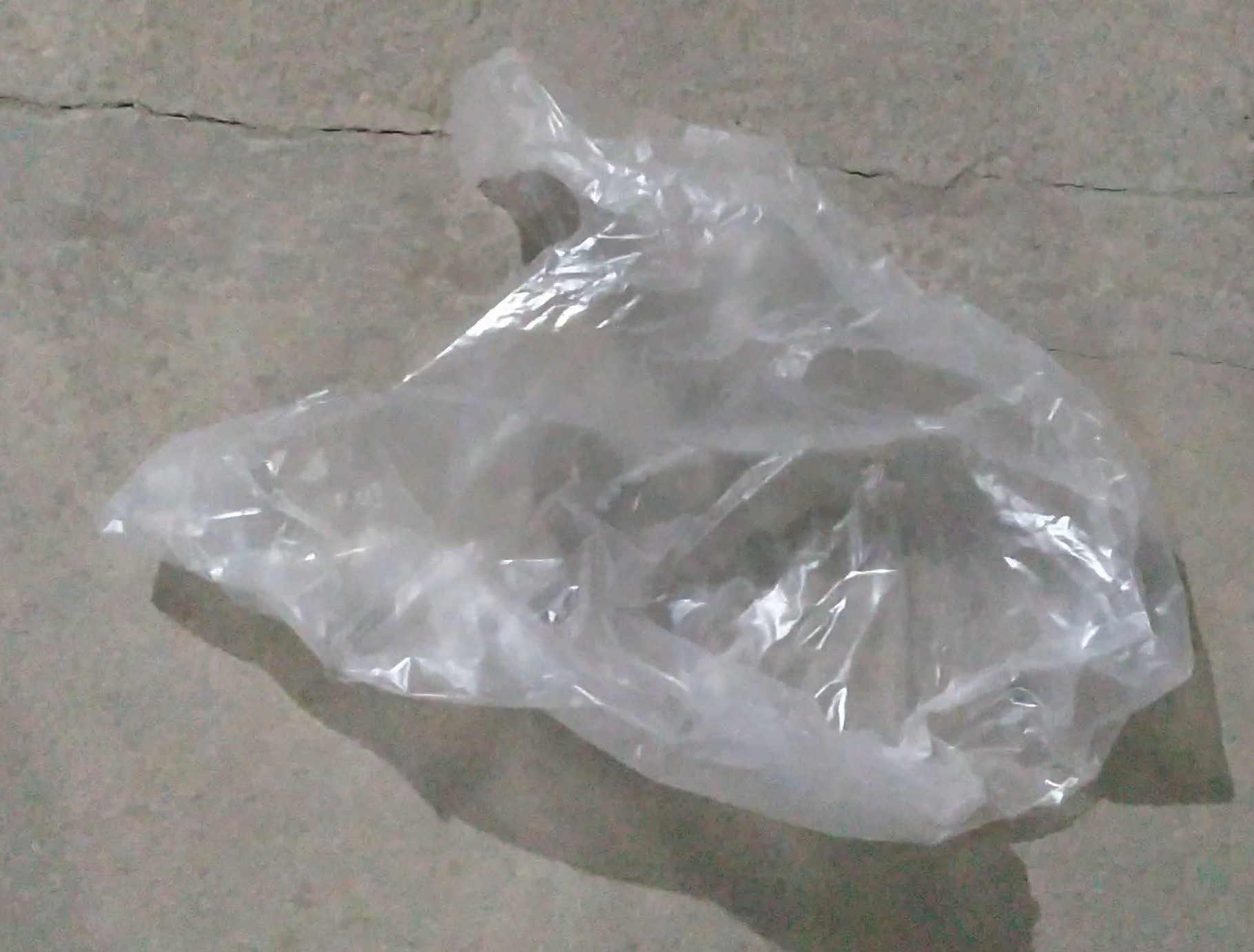}
        \includegraphics[width=4.6cm]{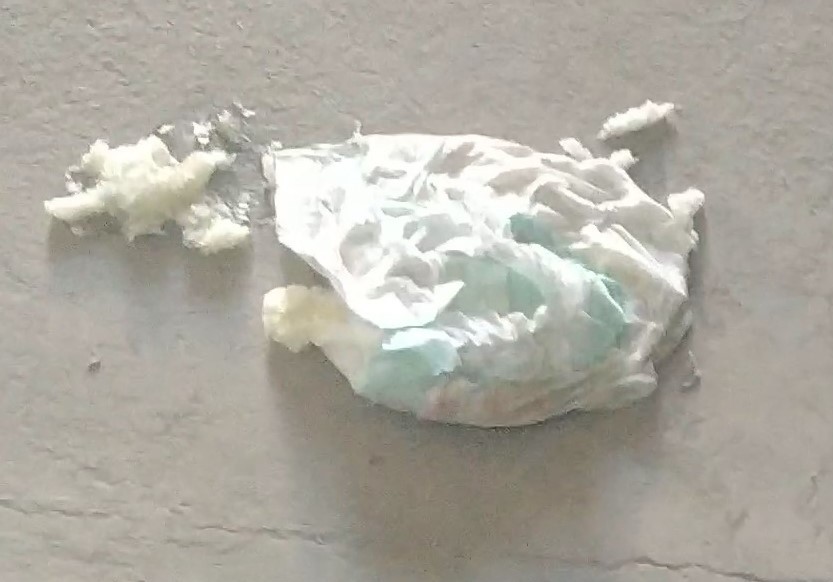}
        \includegraphics[width=4cm]{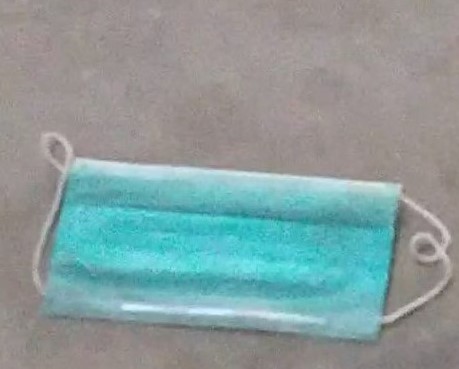}
    \end{center}
    \begin{center}
        Fig 02: Created dataset and sample input images
    \end{center}
\end{center}

\subsection{Annotation {\&} Training Dataset}
The most challenging task was to label the dataset, which was performed using the label image tool; the dataset of 2100 images was labeled for the CNN model. After dividing the video into frames, the labeled dataset was trained on the CNN model. The model performance was measured by accuracy and average weight loss of the model, which is around 0.0930, and the accuracy of the model was 97.2{\%}. To get more accurate and improved results, we add layers in the CNN model. Thus, the results of the improved model justify the detection with its accuracy of 99.6{\%} with average weights loses 0.6928, which satisfies the results. After the detection of the trash, the CNN returned bounding box. Pixels values denote the position of the bounding box of the object; (xmin,ymax), (xmax,ymax), (xmin,ymin), (xmin,ymax)[1].
\newline
\begin{equation}
\left(x_{o}, y_{o}\right)=\left(\frac{x_{\min }+x_{\max }}{2}, \frac{y_{\min }+y_{\max }}{2}\right)
\end{equation}
\begin{equation}
\begin{array}{l}\left(x_{i}, y_{i}\right)=\left(\frac{i m g_{w i d t h}-i m g_{w i d t h}}{2}, \frac{i m g_{h e i g h t}-i m g_{h e i g h t}}{2}\right) \end{array}
\end{equation}
\begin{equation}
\left. (a_{x}(s)\right) =\left(x_{0}-x_{i}\right)=\left(x_{0}\right)
\end{equation}
\begin{equation}
\left. (a_{y}(s)\right) =\left(y_{0}-y_{i}\right)=\left(y_{0}\right)
\end{equation}
\newline
From the above equation, it can be concluded that ax(s) and ay(s) the values must be equal to zero or always approximate for appropriately detecting the object. Furthermore, for successfully detecting, the middle must be similar to the center of the image to detect properly. Middle of the classes detected the bounding box value, which was additionally used for detection and following the as shown in Fig.6. The CCTV camera is used for the CNN model to detecting the trash and identifying the object. The Logitech C310, 720p HD for video recording, captures the image with 1080p/30fps; however, the unique feature is auto-focus with a lens of Full HD glass, and the field of view is 78 degrees, which results in the high-resolution HD video with images.
        \begin{table}[h]
        \begin{center}

        \renewcommand{\arraystretch}{1.0}
        \textbf{Table 2 Parameters and values of the improved CNN model.}
        \resizebox{8.5cm}{!}{
        \begin{tabular}{|c|c|}
        \hline
        \textbf{Parameters} & \textbf{Values}                                                \\ \hline
        Momentum      			   & 0.9                                                                    \\ \hline
        Iterations (t)              & 10,000
                                                                          \\ \hline
        Optimizer                   & \begin{tabular}[c]{@{}c@{}}Stochastic gradient \\ descent (SGD)\end{tabular} \\ \hline
        Batch size                    & 32                                                                          \\ \hline
        Subdivision                   & 16                                                                           \\ \hline
        Learning rate               & 0.001                                                                            \\ \hline
        
         Exposure           & 1.5    
                                        \\ \hline
       
        Filter           & 65
                                        \\ \hline
        Stride           & 1                                            
                                        \\ \hline
        Hue           & .1                                                
                                        \\ \hline
         Saturation           & 1.5                                               
                                        \\ \hline
        Decay          & 0.0005                                \\ \hline  
        Channels          & 3                                           \\ \hline  
  \end{tabular}
 }
 \end{center}
\end{table}

\subsection{Functionality of Model}
In the operational stage, the CNN model was improved by adding the layers which predict more accurate results. Whereas a CCTV camera continuously monitors to detect any trash violations. To detect the image from the model, each video frame is passed through to the CNN. All the detected objects (person) in the given
picture with bounding boxes return from the CNN-based model. Fig 6 illustrates the test images, where objects are detected and assigned boundary boxes. The model is functioning in real-time through a CCTV camera. When the model detecting trash, it makes the video when identifying the object. It records for 10 seconds and saves the video, images in the specified path given to the model.
The new weights are created after 1000 iterations. The weights and highest mean average precision (mAP)  are measured for training. The training was completed on the improved CNN model and acheive results, the mAP of highest achieved, was found to be 99.6{\%} as demonstrated  in the chart of real-time training in Fig 4(b). CNN default accuracy is 89{\%} and its mean average precision (mAP) was 97.2{\%}; however, a decent improvement after adding layers and improved model accuracy was 99.6{\%}. The momentum of 0.9 in default and improved models, a learning rate of 0.001; and other added values are provided in Table 2.Therefore, it has been shown that making a model stronger and deeper to improve its accuracy by changing the function of the model. To achieve best accuracy results, 10000 iterations was done to improved the model  for training, and when the iteration of 1000 was completed, weight file results were built time by time. Calculating the model accuracy, in the testing phase best weight files which consist of highest mAP was further used. The GPU GTX 1660 Super X  was used to trained CNN model, with a batch size of 64, subdivision 8, and filter 65. The improved model detected the object in 3.568.84 ms, moreover, the default model detected the object in 7.895 ms, because the advanced model is much deeper than the default model.
\begin{center}
    \begin{center}
         \includegraphics[width=3.5cm]{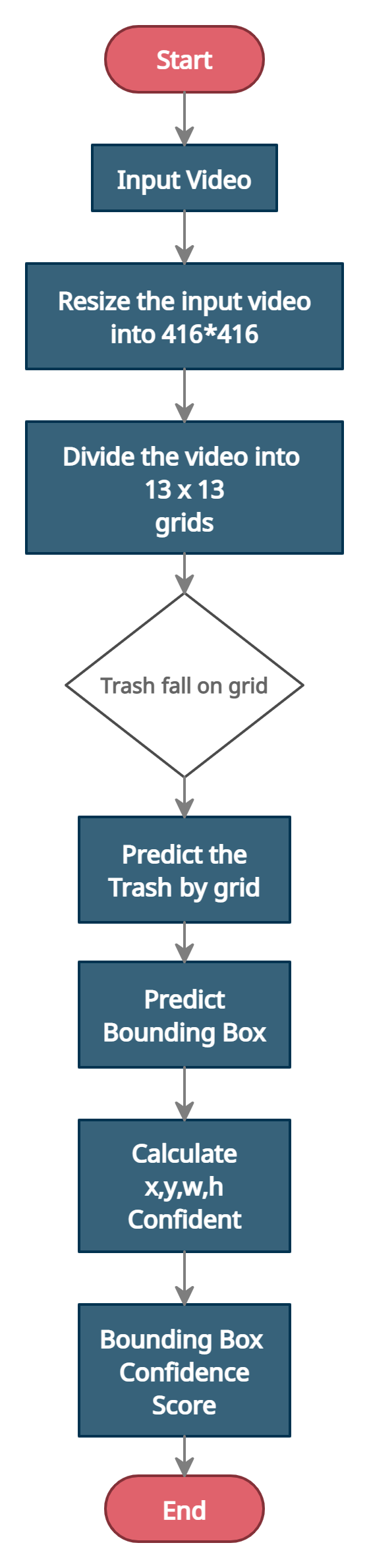}
    \end{center}
   
    Fig 03: Flow chart of Convolutional Neural Network Model.
\end{center}
\begin{center}
    \begin{center}
        \includegraphics[width=8cm]{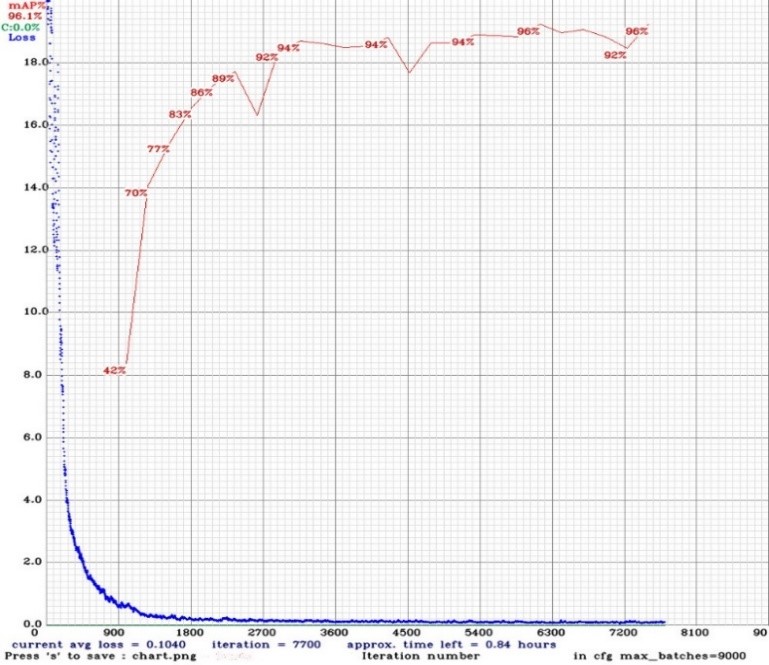}
        \includegraphics[width=8cm]{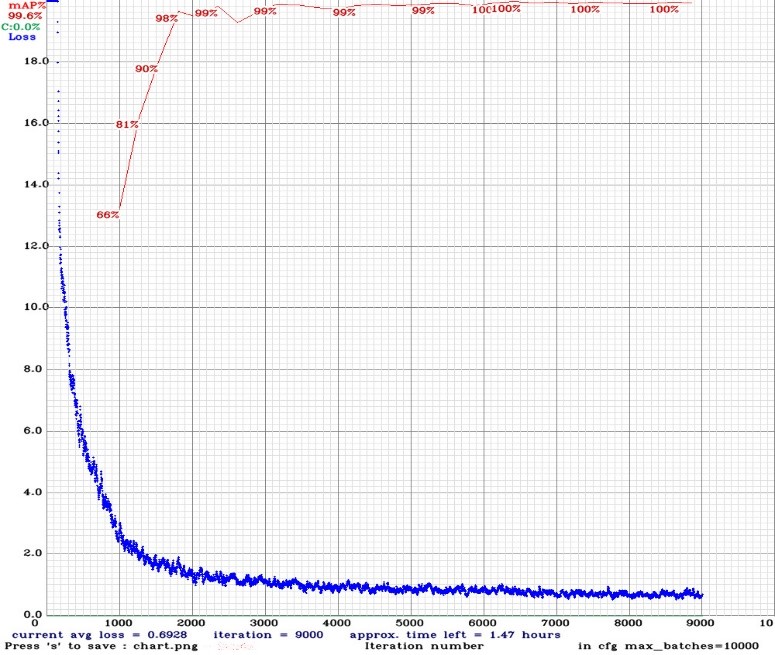}
    \end{center}
    \begin{center}
         Fig 04 (a): Training default deep CNN model.
         Fig 04 (b): Training improved deep CNN model.
    \end{center}
   
\end{center}
\begin{table}[]
    \begin{center}
	\renewcommand{\arraystretch}{1.4}
	\textbf{Table 3 Calculations and differentiation of both models of the CNN. }
	\newline
	\centering
	\resizebox{8cm}{!}{
		\begin{tabular}{|c|c|c|} 
			\hline
			\textbf{Model} & \textbf{\begin{tabular}[c]{@{}c@{}}Mean Average Precision (mAP)\%\\ \end{tabular}} & \textbf{\begin{tabular}[c]{@{}c@{}}Accuracy\%\\ \end{tabular}} \\ \hline
			\textbf{CNN Improved Model}              &       0.0930                                                                                                                                                                                                                                 & 99.6                                                                                  \\ \hline
			\textbf{CNN Default Model} & 0.6928                                                                              & 97.2                                                                                                                                                        \\ \hline
	\end{tabular}}
	    
    \end{center}
\end{table}
\subsection{Evaluation of performance metrics}
Metrics are used to examine the detection of trash, the following parameters of metrics are defined below.

\paragraph{True Positive} 
One of the multiple true output detection is counted as true positive  which occurred within the frame. If trash falls in a defined center place then it is classified as true output detection.
\paragraph{True Negative (TN)} 
The frames without defined trash are called true negative which means that frame is negative but detection is true.

\paragraph{False Positive (FP)}
If detected trash does not fall inside the class ground truth which is defined center place.      

\paragraph{False Negative (FN)} 
The False negative defined as, trash in the class are missing in the frame.

\paragraph{Precision}
The precision is defined as, improved model detects defined objects in a class within an image which the metric shows calculation was precisely.
\begin{equation}
\text {Precision }(\text {Pre})=\frac{T P}{T P+F P} \times 100
\end{equation}

\paragraph{Sensitivity}
The sensitivity metrics is known as recall, or true positive rate, real class of the defined object placed correctly which measures the proportion.
\begin{equation}
\text {Sensitivity }(\text {Sen})=\frac{T P}{T P+F N} \times 100
\end{equation}

\textbf{F1- score and F2- score:} The F1 and F2 score is convey the balance between the precision and recall, it is also known as harmonic mean and limit is in between 0,1.
\begin{equation}
F 1-\text {score}=\frac{2 \times \text {Sen } \times \text { Pre}}{\text {Sen }+\text {Pre}} \times 100
\end{equation}
\begin{equation}
F 2- score=\frac{5 \times \text {Pre} \times \text {Sen}}{4 \times \text {Pre}+\text {Sen}}
\end{equation}

\begin{center}
    \begin{center}
        \includegraphics[width=16cm]{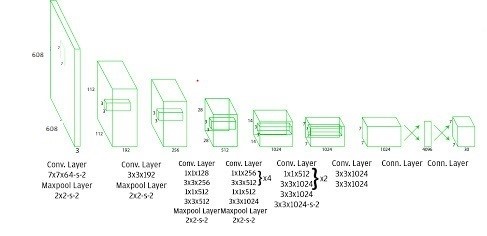}
        \begin{center}
            Fig 05: Explanation of architecture Convolutional Neural Network (CNN) Model
        \end{center}
    \end{center}
\end{center}
\subsection{Loss Function}
The network of  CNN has superficial addition distinction, such as IOU errors, classifications errors, and coordinates errors. The process of the CNN uses the loss calculation noted as a sum-square error.  The loss function formula is given below.
\begin{equation}
loss = \sum\limits_{t = 0}^{{s^2}} {coordErr}  + iouErr + clsErr 
\end{equation}
To estimate overall loss function the weight of each loss function is calculated. The model exhibits unstable behavior and dissimilarity during the training phase when the classification error is constant with a coordinate error. Therefore, the coordinate error weight was fixed to $\lambda$ = 5. The CNN employs  $\lambda$noobj for the IOU error to avoid confusion between the no object grid and object grid. While training dataset the absolute loss function obtained can be described as follows.
\begin{equation}
\begin{array}{l}
loss = \lambda _{coord}^{}\sum\limits_{t = 0}^{s2} {} \sum\limits_{j = 0}^B {} l_{ij}^{obj}\left[ {{{\left( {{x_i} - {{\hat x}_i}} \right)}^2} + {{\left( {{y_i} - {{\hat y}_i}} \right)}^2}} \right]\\
 + {\lambda _{coord}}\sum\limits_{t = 0}^{s2} {} \sum\limits_{j = 0}^B {} l_{ij}^{obj}\\
 \left[ {{{\left( {\sqrt {{w_i}}  - \sqrt {{{\hat w}_i}} } \right)}^2} + {{\left( {\sqrt {{h_i}}  - \sqrt {{{\hat h}_i}} } \right)}^2}} \right]\\
 + \sum\limits_{t = 0}^{s2} {} \sum\limits_{j = 0}^B {} l_{ij}^{obj}{\left( {{c_i} - {{\hat c}_i}} \right)^2}\\
 + {\lambda _{noobj}}\sum\limits_{i = 0}^{g2} {} \sum\limits_{j = 0}^B {} l_{ij}^{obj}{\left( {{c_i} - {{\hat c}_i}} \right)^2}\\
 + \sum\limits_{i = 0}^{g2} {} l_i^{obj}\sum\limits_{c \in class}^{} {} {\left( {{p_i}(c) - {{\hat p}_i}(c)} \right)^2}
\end{array}
\end{equation}
The above equation as demonstrated each of the cell numbers corresponding to the prediction boxes is indicated by B and the number of grids is represented by g. 
The coordinate center of each cell is defined as (x, y); respectively its width and height are indicated as (h, w). Moreover, the loss function position of the weight is represented as  $\lambda$coord and the prediction box confidence is shown as C; the confidence of objects in the class is labeled as R. The loss function weight is classified as  $\lambda$noobj. The objects trained in this class if present, the value is set as 1, and otherwise, it is 0 [3].s
\begin{center}
\begin{center}
    \includegraphics[width=4cm]{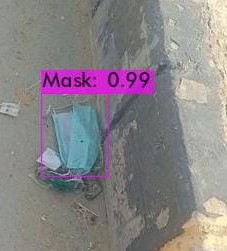}     \includegraphics[width=4.1cm]{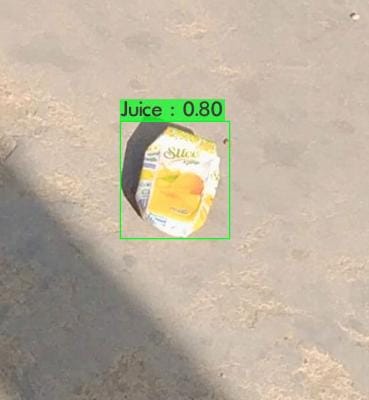}
    \includegraphics[width=4.6cm]{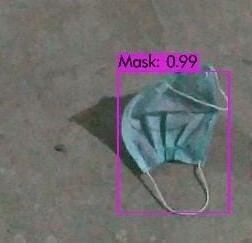}
\end{center}
\begin{center}
    \includegraphics[width=3cm]{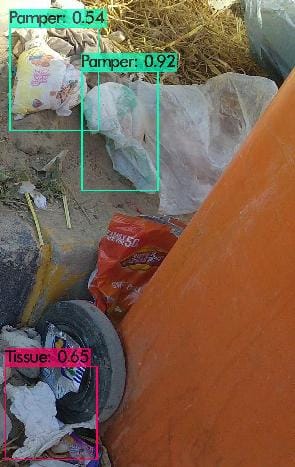}
    \includegraphics[width=5.3cm]{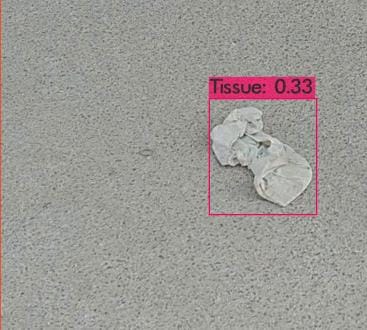}
    \includegraphics[width=5.3cm]{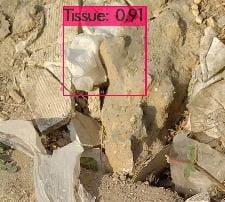}    
\end{center}
\begin{center}
    \includegraphics[width=3.4cm]{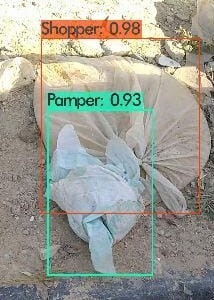}
    \includegraphics[width=5.4cm]{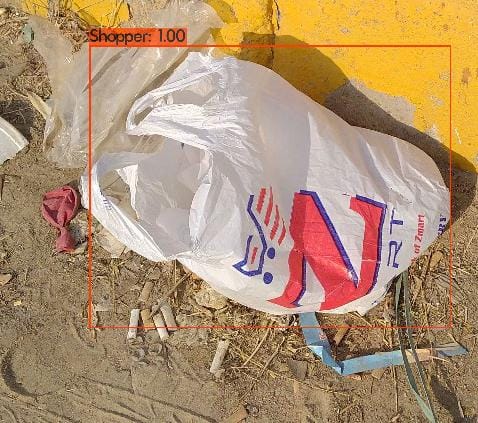}
    \includegraphics[width=3.8cm]{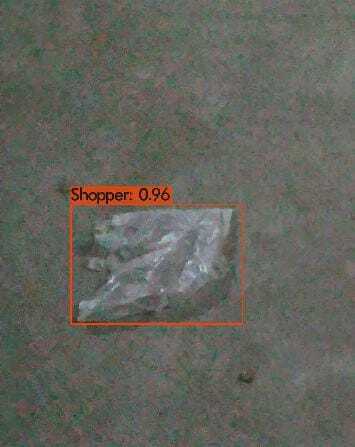}
\end{center}

\begin{center}
    Fig 06: Output results of detected trash through CCTV.
\end{center}
    
\end{center}

\subsection{CONCLUSION AND FUTURE WORK}
Concluding all the highlighted points, a  model of deep convolutional neural network approach was used for real-time trash detection using a CCTV camera identifying an object. After, a time taken struggle is made to create a custom data set due to the unavailability of specific trash classes. Similarly, the evaluated results for both trained models with specific datasets are given in time, mAP and accuracy.
\newline
The future work, for this paper, can be done using a larger number of a dataset. Thermal CCTV cameras can be used for detecting trash with more classes in future work.

\section*{Acknowledgment}
This form is about making conformation for the acknowledge of the submitted manuscript, Real-Time Trash Detection for Modern Societies using CCTV to Identifying Trash by utilizing Deep Convolutional Neural Network, in TURKISH JOURNAL OF ELECTRICAL ENGINEERING {\&} COMPUTER SCIENCES. All authors have participated in (a) conception and design, or analysis and interpretation of the data (b) drafting the article or revising it critically for important intellectual content and (c) approval of the final version. This manuscript has not been submitted to, nor is under review at, another journal or other publishing venue. The authors have no conflict of interest whatsoever in submitting the manuscript in the particular journal.
\textbf{
\newline
1st Author name: Syed Muhammad Raza and author contributions, e.g., as follow: Introduction, Related work, Experimental work, Improved Model,  and wrote paper.
\newline
Affiliation: PAF-Karachi Institute of Economics Technology, Pakistan 
\newline
Dated: 2021.09.06
\newline
2nd Author name: Syed Muhammad Ghazi and author contributions, e.g., as follow: Diagrams,Tables, Calculation, Equations, reference work and wrote paper.
\newline
Affiliation: Kumoh National Institute of Technology, South Korea 
\newline
Dated: 2021.09.06
\newline
3rd Author name: Syed Ali Hassan and author contributions, e.g., as follow: gave idea, and gave structure of work to perform and review paper.
\newline
Affiliation: Kumoh National Institute of Technology, South Korea
\newline
Dated: 2021.09.06
\newline
Corresponding Author name: Dr. Soo Young Shin and author contributions, e.g., as follow: review paper, correction work, finalize the paper. 
\newline
Affiliation: Kumoh National Institute of Technology,South Korea 
\newline
Dated: 2021.09.06
}

\end{document}